\documentclass[runningheads]{llncs}
\usepackage[T1]{fontenc}
\usepackage{graphicx}
\usepackage{hyperref} 
\usepackage{multirow}
\usepackage{color}

\begin{document}
\title{Intrapartum Ultrasound Image Segmentation of Pubic Symphysis and Fetal Head Using Dual Student-Teacher Framework with CNN-ViT Collaborative Learning}
\titlerunning{Dual Student-Teacher Combining CNN and Transformer}
\author{Jianmei Jiang\inst{1} % 1{Jiang, Jianmei}
\and
Huijin Wang\inst{1}  % 2{Wang, Huijin}
\and
Jieyun Bai\inst{1,2,3} \thanks{Jieyun Bai: \email{jbai996@aucklanduni.ac.nz}} % 3{Bai, Jieyun}
\and
Shun Long\inst{1} % 4{Long, Shun}
\and
Shuangping Chen\inst{1} % 5{Chen, Shuangping}
\and
Victor M. Campello \inst{4,5} % 6{Campello, Victor M.}
\and
Karim Lekadir\inst{4,5} % 6{Lekadir, Karim}
}
\authorrunning{J. Jiang et al.}
\institute{College of Information Science and Technology, Jinan University, Guangzhou 510632, China \and
%\email{jbai996@aucklanduni.ac.nz}\and
Guangdong Provincial Key Laboratory of Traditional Chinese Medicine Information Technology, Jinan University, Guangzhou 510632, China\and
Auckland Bioengineering Institute, University of Auckland, Auckland 1010, New Zealand\and
Department de Matemàtiques i Informàtica, Universitat de Barcelona, Barcelona 08007, Spain\and
Institució Catalana de Recerca i Estudis Avançats (ICREA), Barcelona 08007, Spain
}
\maketitle              % typeset the header of the contribution
\begin{abstract}
The segmentation of the pubic symphysis and fetal head (PSFH) constitutes a pivotal step in monitoring labor progression and identifying potential delivery complications. Despite the advances in deep learning, the lack of annotated medical images hinders the training of segmentation. Traditional semi-supervised learning approaches primarily utilize a unified network model based on Convolutional Neural Networks (CNNs) and apply consistency regularization to mitigate the reliance on extensive annotated data. However, these methods often fall short in capturing the discriminative features of unlabeled data and in delineating the long-range dependencies inherent in the ambiguous boundaries of PSFH within ultrasound images. To address these limitations, we introduce a novel framework, the Dual-Student and Teacher Combining CNN and Transformer (DSTCT), which synergistically integrates the capabilities of CNNs and Transformers. Our framework comprises a Vision Transformer (ViT) as the 'teacher' and two 'student' models — one ViT and one CNN. This dual-student setup enables mutual supervision through the generation of both hard and soft pseudo-labels, with the consistency in their predictions being refined by minimizing the classifier determinacy discrepancy. The teacher model further reinforces learning within this architecture through the imposition of consistency regularization constraints. To augment the generalization abilities of our approach, we employ a blend of data and model perturbation techniques. Comprehensive evaluations on the benchmark dataset of the PSFH Segmentation Grand Challenge at MICCAI 2023 demonstrate our DSTCT framework outperformed 10 contemporary semi-supervised segmentation methods. Code available at https://github.com/jjm1589/DSTCT.
\end{abstract}
\section{Introduction}
The segmentation of the pubic symphysis and fetal head (PSFH) from intrapartum ultrasound (US) images is a critical step in developing an automated diagnostic system. This process is essential for generating quantitative descriptors, such as shape and size of PSFH. These descriptors are crucial in assessing labor progression and identifying potential delivery complications \cite{chen2024direction,angeli2022automatic,ou2024rtseg,bai2024new,qiu2024psfhsp,chen2024psfhs,chen2024fetal,zhou2023segmentation}. Recent advancements in deep learning, particularly in Convolutional Neural Networks (CNNs) and Transformers, have significantly improved medical image segmentation. However, these models face challenges in clinical applications due to the scarcity of large-scale annotated training datasets. Deep learning approaches generally require extensive, labeled datasets to ensure model generalization. The collection of densely annotated US images is a complex task. It demands considerable time, medical expertise, and clinical experience for accurate pixel-wise labeling \cite{bai2022framework,zhou2020automatic}. In clinical practice, there is often a larger amount of unlabeled data than labeled data, emphasizing the importance of semi-supervised learning techniques. These techniques aim to improve the segmentation performance of US images by leveraging unlabeled data, a practice that is garnering increasing research interest.

Despite the progress in semi-supervised learning for US image segmentation, challenges persist due to common US imaging issues like shadow artifacts and unclear boundary lines. Most semi-supervised approaches rely on CNN architectures, which may lead to under-segmentation or over-segmentation due to their localized processing nature. In contrast, Transformer-based models offer a promising alternative. They excel in capturing wide-ranging, non-local interactions, potentially resolving CNNs' inherent limitations. These models are adept at identifying and assimilating features from distant but visually similar regions, thus improving network feature discrimination. However, their effectiveness largely hinges on having access to extensive annotated datasets—a condition seldom met in medical imaging due to the limited availability of labeled data. Thus, developing methods to efficiently train Transformer-based models with a limited set of annotated data remains a formidable challenge in the area of US image segmentation.

This paper proposes a novel framework, the Dual-Student and Teacher Combining CNN and Transformer (DSTCT). Our approach uses a dual-student configuration, employing cross-supervision with hard pseudo labels to expand the training dataset. Additionally, our Consistency Learning with Soft Pseudo Labels (CLS) strategy aims to mitigate label noise and foster entropy minimization. Given the stark differences between CNN and Transformer models, we introduce the Classifier Deterministic Difference (CDD) \cite{li2021bi} to harmonize predictions between the two model types. The teacher model applies consistency regularization (CR) constraints, guiding the dual-student networks. This involves using data and model perturbations to align the dual-student networks' prediction maps with those of the teacher model, thus bolstering the overall model generalization. The main contributions of our work include: (1) we present the Dual-Student and Teacher Combining CNN and Transformer (DSTCT) framework. This new approach significantly enhances the utility of unlabeled data in semi-supervised segmentation of ultrasound (US) images; (2) our proposed framework effectively integrates hard and soft pseudo-label learning, entropy minimization, and consistency regularization; (3) through rigorous quantitative analysis, we establish that our DSTCT framework surpasses the performance of ten existing semi-supervised techniques in accurately segmenting the pubic symphysis and fetal head (PSFH) in intrapartum ultrasound images.

\section{Method}
For general semi-supervised learning, the training set is composed of a labeled dataset $D^l_N$  with $N$ labeled images and an unlabeled dataset $D^u_M$ with $M$ ($M >> N$) raw images, and the full training dataset is denoted as $D_{N+M}$ = $D^l_N \cup D^u_M$. For an image $X \in D^l_N$, its ground truth $Y$ is available. But the ground truth is not provided for $X \in D^u_M$. $P_{s1}$ and $P_{s2}$ are the probability outputs derived from student1 and student2, respectively. $P^*_{s1}$ is student1's soft pseudo labels.
\begin{figure}
\centering
\includegraphics[width=\textwidth]{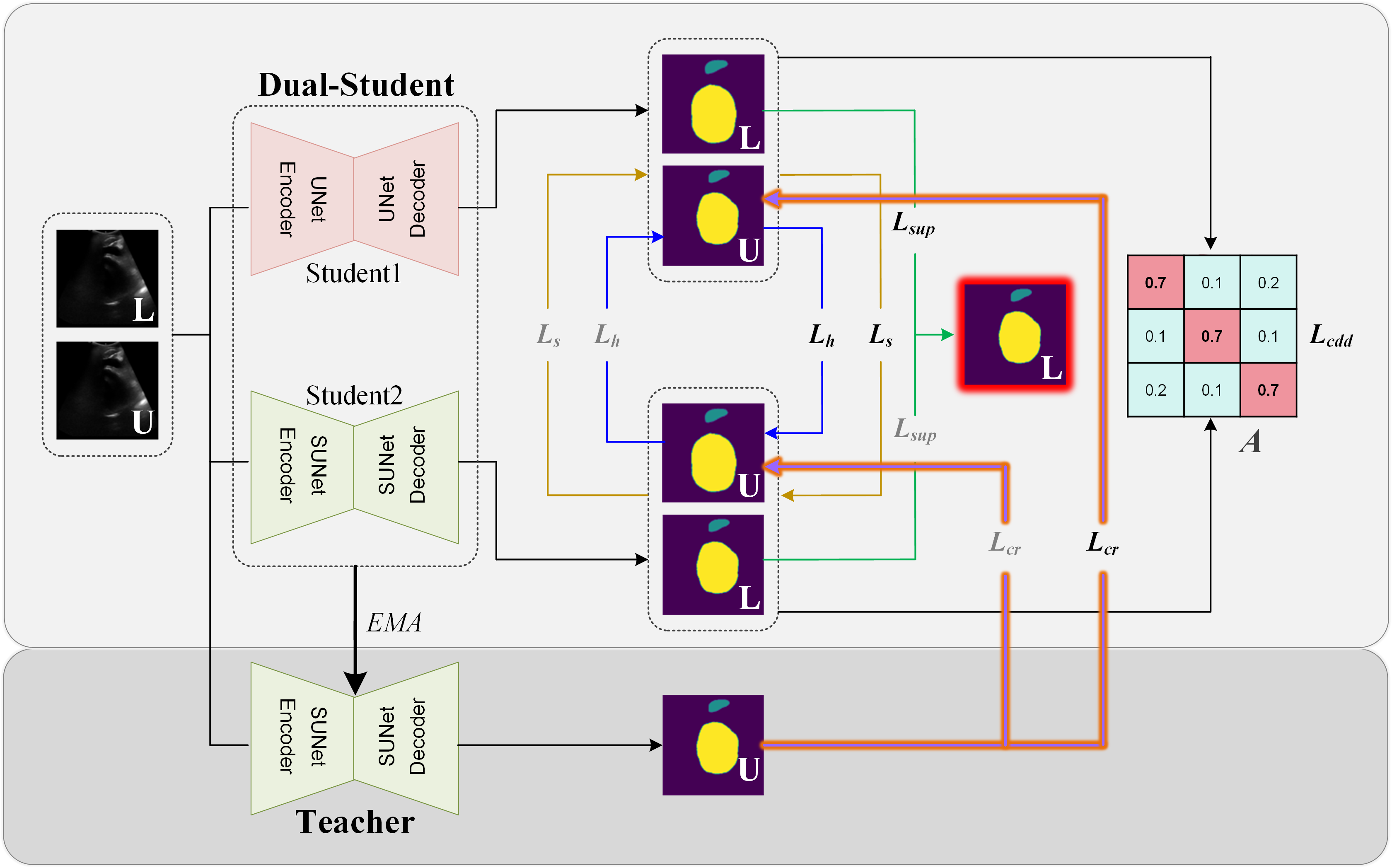}
\caption{An overview of the DSTCT architecture, where the black and grey parts are the loss functions for student1 and student2, respectively} \label{fig1}
\end{figure}

An overview of the proposed DSTCT architecture is shown in Fig.\ref{fig1}. It comprises a dual-student model and a single-teacher model. The dual-student model consists of a CNN model UNet (namely student1) and a Transformer model Swin-UNet (SUNet, namely student2), while the single-teacher model uses the same Transformer model SUNet. In particular, our DSTCT tackles the semi-supervised image segmentation from five various aspects: supervised learning ($L_{sup}$), cross-supervision with hard pseudo labels ($L_h$), consistency learning with soft pseudo labels ($L_s$), minimization of classifier determinacy discrepancy ($L_{cdd}$) and consistency regularization constraints from the teacher model ($L_{cr}$). Therefore, the overall training loss function for student1 or student2 can be defined as:
\begin{equation}
    L_{total} = L_{sup} + \alpha L_h + \beta L_s + \gamma L_{cdd} + \mu L_{cr}
\end{equation}
where $\alpha$, $\beta$, $\gamma$ and $\mu$ are trade-off weights, which were set $\alpha$ = 0.5, $\beta$ = 1.0, $\gamma$ = 3.0, and $\mu = 0.1$ \cite{wang2022cnn} , respectively, in the proposed cooperative training process. Note: $L_{cdd}$ is a common part of both students.\\
\textbf{Supervised learning ($L_{sup}$).} We use the labeled data to train the student models. Cross-entropy loss $L_{ce}$ and Dice loss $L_{dice}$ are used as follows:
\begin{equation}
    L_{sup} = \frac{1}{2} \sum_{X \in D^l_N}(L_{ce}(P_{s1},Y) + L_{dice}(P_{s1},Y)) 
\end{equation}
\textbf{Cross-supervision with hard pseudo labels ($L_h$).} The predictions between the CNN and ViT have different properties, essentially in the output level. Based on their predictions, we employ $argmax(.)$ function to yield the hard pseudo labels for the cross-supervision between the peer networks. The cross-supervision loss of unlabeled data is defined as:
\begin{equation}
    L_{h} = \sum_{X \in D^u_M}L_{dice}(P_{s1},argmax(P_{s2}))
\end{equation}
\textbf{Consistency learning with soft pseudo labels ($L_s$).} However, the hard pseudo labels generated by the maximum confidence are inevitably noisy, which may cause confusion bias during the segmentation training. To further reduce the noise of the hard pseudo labels and focus on unlabeled challenging regions, a sharpening function \cite{li2022collaborative,xie2020unsupervised} is utilized to generate soft pseudo labels, which can decrease the prediction uncertainty of the models. Soft pseudo labels can be obtained as follows:
\begin{equation}
   P^*_{s1} = \frac{(P_{s2})^{1/\tau}}{(P_{s2})^{1/\tau} + (1 - P_{s2})^{1/\tau}} 
\end{equation}
where $\tau$ is a hyper-parameter to control the temperature of sharpening, and is set to 0.1 in our experiment \cite{Wugzxc22}. Consistency learning is performed between the probability output of one model and the soft pseudo label of the other. The final loss function of consistency learning with soft pseudo labels is:
\begin{equation}
    L_{s} = \sum_{X \in D_{N+M}}E[P_{s1},P^*_{s1}]
\end{equation}
where $E$ is the Mean Squared Error (MSE) loss. \\
\textbf{Classifier Determinacy Disparity ($L_{cdd}$).} The diversity caused by structural differences between CNN and ViT may compromise the model's performance. Directly aligning the discrepancy between the predictions is somewhat not confident. Therefore, we investigate classifier discrepancy by Bi-classifier Prediction Relevance Matrix: $A = P_{s1} {P_{s2}}^T$ \cite{li2021bi}. The sum of the diagonal elements in matrix $A$ represents the consistency of the classifier's predictions, while the off-diagonal elements reflect the uncertainty of the predictions. We aim to maximize the former and minimize the latter. The minimizing classifier determinacy discrepancy loss is defined as follows:
\begin{equation}
    L_{cdd} = \sum_{X \in D_{N+M}} [\sum_{m,n=1}^{C}A^{m,n} - \sum_{m=1}^{C}A^{m,m}]
\end{equation}
where $A^{m,n}$ denotes the element in the $m$-th row and $n$-th column, and $C$ is the number of categories. \\ 
\textbf{Consistency Regularization ($L_{cr}$).} The teacher architecture aims to minimize discrepancies between the predictions of the dual-student networks and the teacher network under both data and network perturbations. The consistency loss between the output probabilities of them is defined as follows:
\begin{equation}
    L_{cr} = \sum_{X \in D_M}E [f_t(X;\bar{\theta};\sigma^{'}),f_{s1}(X;\varphi;\sigma)]
\end{equation}
where $\sigma$ and $\sigma^{'}$ represent different data perturbations and random dropout operations at the network layer, and $\varphi$ and $\bar{\theta}$ represent the network parameter of student1 and teacher. $\bar{\theta}$ is updated via exponential moving average (EMA) from student2's network. $E$ is the MSE loss. 
\section{Experiments and Results}
\subsection{Dataset and Implementations} 
\textbf{Dataset.} This study leverages a dataset sourced from the MICCAI 2023 Grand Challenge \cite{lu2022jnu}, focused on the segmentation of the pubic symphysis (PS) and fetal head (FH). The comprehensive dataset includes 5,101 images, which have been methodically partitioned into training (70\%), validation (10\%), and testing (20\%) subsets. Each image is paired with a precise segmentation mask delineating the FH and the PS, enabling the effective development and evaluation of pertinent segmentation models. \\
\textbf{Implementation details.} In this study, a compact version of ViT, pre-initialized with ImageNet weights is used for Swin-UNet (SUNet) architecture. The implementation was implemented on an Ubuntu 20.04 operating system, utilizing Python 3.8, PyTorch 1.10, and CUDA 11.3. A Tesla T4 GPU facilitated the computations. The network training regimen encompassed 30,000 iterations, employing the Stochastic Gradient Descent (SGD) optimizer, configured with a momentum of 0.9 and a weight decay of 0.0001. We adopted a batch size 16, comprising an equal split of eight labeled and eight unlabeled images to support semi-supervised learning paradigms. The training commenced with an initial learning rate of 0.01, which was dynamically adjusted using a clustering-based learning rate strategy. Additionally, we introduced data noise perturbation within the range of [-0.2, 0.2] to enhance model robustness against input variability. \\
\textbf{Comparison Strategy.} To verify the effectiveness of our DSTCT, we compare it with full/limited supervision (FS/LS) baselines and several state-of-the-art approaches for semi-supervised segmentation, including mean teacher (MT) \cite{tarvainen2017mean}, deep adversarial network (DAN) \cite{zhang2017deep}, uncertainty-aware mean teacher (UAMT) \cite{Yuwxfh19}, cross consistency training (CCT) \cite{ouali2020semi}, cross pseudo-supervision (CPS) \cite{chenyzgw21semi}, cross teaching between CNN and transformer (CTCT) \cite{luohswzs22semi}, interpolation consistency (ICT) \cite{verma2022interpolation}, deep co-training (DCT) \cite{qiao2018deep}, self-integration method based on consent-aware pseudo-labels (S4CVnet) \cite{wang2022cnn}, collaborative Transformer-CNN learning (CTCL) \cite{li2022collaborative}.\\
\textbf{Evaluation Metrics.} The performance of these models was quantitatively evaluated using three established metrics: the Dice Similarity Coefficient (DSC), the 95\% Hausdorff Distance ($HD_{95}$), and the Average Surface Distance (ASD). These metrics facilitate a thorough assessment of the segmentation models' accuracy and consistency, critical for validating their clinical applicability.
\subsection{Comparison with Other Methods} 
Table \ref{tab1} presents the quantitative outcomes on the PSFH dataset when trained with  20\% of the total labeled training data. Notably, our proposed method significantly outperforms existing state-of-the-art approaches across most evaluation metrics. Remarkably, with only 20\% labeled data for training, our DSTCT achieves 89.3\% DSC performance, only 6\% inferior to the upper bound performance.
\begin{table}
\center
\scriptsize
\caption{Quantitative comparison. The best result is in bold.} \label{tab1}
% \scalebox{1.15}{
\begin{tabular}{l l |c c c| c c c |c c c} 
 \hline
 \multirow{2}*{Labeled} & \multirow{2}*{Method} & \multicolumn{3}{c|}{PS} & \multicolumn{3}{c|}{FH} & \multicolumn{3}{c}{PSFH} \\ 
 \cline{3-11}
    & & DSC$\uparrow$   & ASD$\downarrow$   & $HD_{95}\downarrow$  & DSC$\uparrow$   & ASD$\downarrow$   & $HD_{95}\downarrow$  & DSC$\uparrow$   & ASD$\downarrow$   & $HD_{95}\downarrow$    \\
                    
\hline
\multirow{12}*{20\%} & LS(UNet) & 0.835 & 0.793 & 3.875 & 0.897 & 1.901 & 8.868 & 0.866 & 1.347 & 6.372 \\ 
    &MT	& 0.803	& 0.760	&5.143	&0.872	&2.912	&13.608	&0.837&	1.836&	9.376 \\
    &ICT&\underline{0.847}&0.813 &3.806	&0.896	&1.694	&7.191	&0.871&	1.254&	5.498 \\
    &UAMT & 0.830 &0.801 & 4.375 &0.887	&1.661	&7.296	&0.859&	1.231&	5.836 \\
    &DAN&0.824 & 1.048 & 4.583	& 0.886 &1.443	& 6.871 & 0.855&1.246&	5.727 \\
    &DCT & 0.801 & 1.009 & 4.960 &0.879	&2.081	&9.044	&0.840&	1.545&	7.002 \\
    &CCT& 0.824	& \underline{0.596} & 4.388 &0.886 &2.693 &	11.604 &0.855&	1.645&	7.996 \\ 
    &CPS &0.829	&0.701 & \underline{3.788} & 0.883	& 1.385	& 5.952	&0.856&	1.043&	4.870 \\
    &CTCT &0.836 & 0.980 & 4.265 & \underline{0.908} & 1.340 & 6.126 & 0.872&	1.160&	5.196 \\
    &CTCL& 0.826 & 1.197 & 5.509 & 0.900 & \underline{0.747}& 5.107	& 0.863& 0.972 & 5.308 \\
    &S4CVnet &0.838	& 0.823	& 3.962	& 0.906	& 0.975&\underline{4.658} & \underline{0.872} &	\underline{0.899}&	\underline{4.310} \\
    &OURS & \textbf{0.852} & \textbf{0.581} & \textbf{3.331} & \textbf{0.935} & \textbf{0.351}& \textbf{2.150}&\textbf{0.893} & \textbf{0.466}	& \textbf{2.740} \\
\hline
\multirow{2}*{100\%} & FL(UNet) & 0.943 & 0.113 & 1.103 & 0.964 & 0.086 & 0.605 & 0.953 & 0.099 & 0.854 \\ 
    & FL(SUNet) &  0.931 & 0.143 & 1.248 & 0.958 & 0.189 & 0.789 & 0.944 & 0.166 & 1.018 \\
 \hline
\end{tabular}
\end{table}

\begin{figure} 
\includegraphics[width=\textwidth]{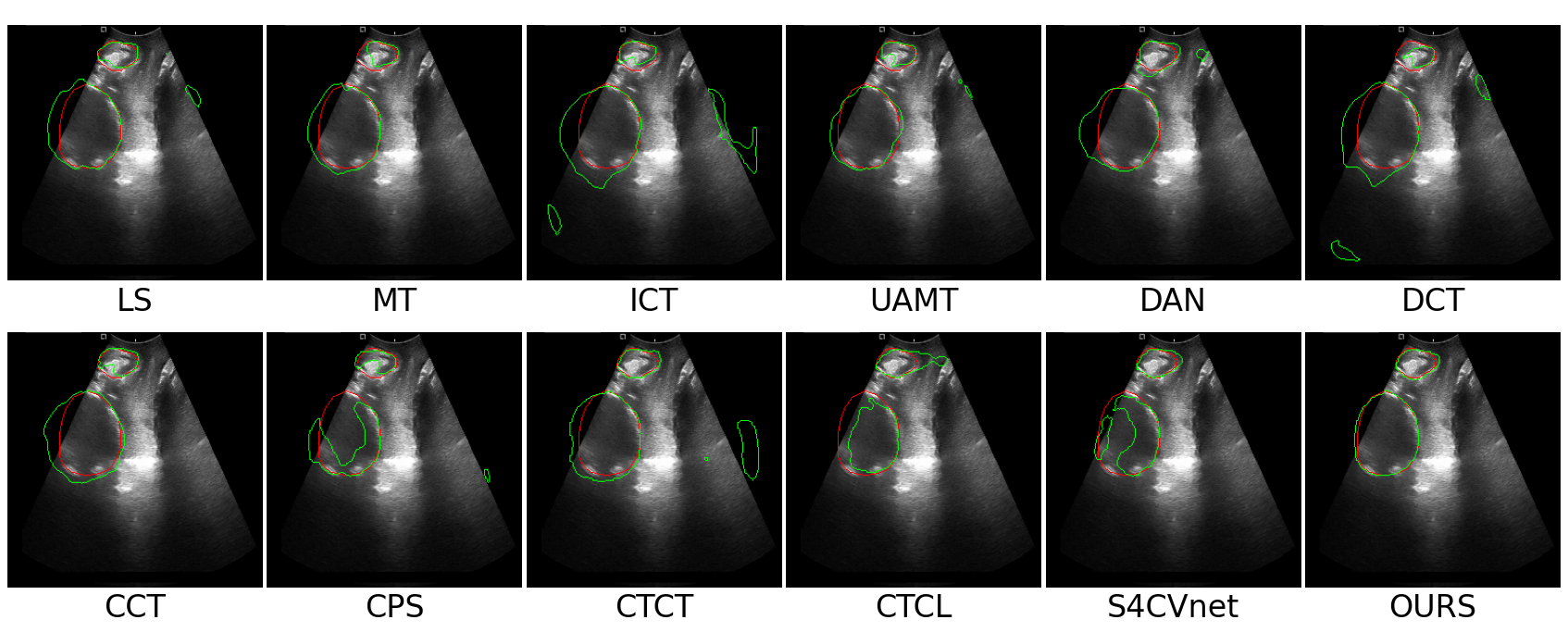}
\caption{Visual comparison of different methods when using 20\% labeled data for testing. The figure shows the ground truth in red and the predicted results in green.} \label{fig2}
\end{figure}

Additionally, Fig.\ref{fig2} offers visual comparisons between the segmentation results. These visualizations highlight that our method achieves more accurate segmentation with reduced false positives and fewer missed segmentation areas. A detailed examination of the comparative experiment results reveals that our proposed framework effectively integrates classical semi-supervised learning strategies, including pseudo-label learning, entropy minimization, and consistency regularization. This integration underscores the efficacy of our approach in leveraging semi-supervised learning techniques for improved segmentation performance.
\subsection{Ablation Study}
\textbf{Training strategy analysis.} Each component of our DSTCT contributes differently to the enhancement of semi-supervised learning. From Table \ref{tab2}, it can be seen that each of the components and the combination of them can improve the PSFH segmentation performance, thus demonstrating the effectiveness of our method. Specifically, CDD and CR improve the DSC performance by 0.8\% and 0.6\%, respectively. By combining CLS, CDD and CR, the DSC performance is improved by 2.1\%, the ASD is decreased by 0.626 mm, and the $HD_{95}$ is decreased by 3.767 mm.\\
\begin{table}
\center
\caption{Different strategies. Where s1, s2, and t are student1,student2, and teacher, respectively. The initial values of the trade-off weights of the loss function are $\alpha$ = 0.1, $\beta$ = 0.1, $\gamma$ = 0.01, and $\mu$ = 0.1, respectively. The best result is in bold.}\label{tab2}
\begin{tabular} {p{1cm} p{1cm} p{1cm}| c c c |c c c |c c c} 
\hline
 \multicolumn{3}{c|}{Designs} & \multicolumn{3}{c|}{PS} & \multicolumn{3}{c|}{FH} & \multicolumn{3}{c}{PSFH} \\ 
 \cline{1-12}
    CLS   & CDD   & CR  & DSC$\uparrow$   & ASD$\downarrow$   & $HD_{95}\downarrow$  & DSC$\uparrow$   & ASD$\downarrow$   & $HD_{95}\downarrow$  & DSC$\uparrow$   & ASD$\downarrow$   & $HD_{95}\downarrow$    \\
\hline
   & & &0.836 &	0.688 &3.821 &	0.890 & 2.133 & 9.741 & 0.863 & 1.410 & 6.781 \\
    $\surd$ & & &0.827 & 1.048 &4.413& 0.892	& 1.542	&6.607 &0.860 &1.295 &5.510 \\
    &$\surd$& &0.830 & 1.002 & 4.323 & 0.912 & 0.815 & 3.775 & 0.871 & 0.909 & 4.049 \\
     & &$\surd$ &0.838 & 0.844 & 3.892 & 0.900 & 1.274 & 5.546 & 0.869 & 1.059 & 4.719 \\
    $\surd$ &$\surd$ & &\textbf{0.848} & 0.731 & 3.819 & 0.902 & 1.555 & 2.493 & 0.875 & 1.143  & 5.294 \\
    $\surd$& &$\surd$ &0.835 & 0.881 & 3.985 & 0.914 & 0.775 & 6.769 & 0.874 & 0.828 & 3.748 \\
    & $\surd$&$\surd$&0.842 & 0.606 & \textbf{3.594} & 0.891 & 1.742 & 7.664 & 0.866 & 1.174 & 5.829 \\ 
    $\surd$ &$\surd$ &$\surd$ &0.837	& \textbf{0.598} & 3.734 & \textbf{0.932} & \textbf{0.429} & \textbf{2.295} & \textbf{0.884}	& \textbf{0.514} & \textbf{3.014} \\
  \cline{1-12}
     s1 & s2  &  t &   &   &   &   &    &   &    &    & \\
 \hline
    UNet& SUNet & SUNet & 0.837 & \textbf{0.598} & \textbf{3.734} &\textbf{0.932} & \textbf{0.429} & \textbf{2.295} & \textbf{0.884} & \textbf{0.514} & \textbf{3.014} \\
     SUNet& UNet & UNet & 0.845 & 0.902 & 4.205 & 0.890 & 1.515 & 6.305 & 0.867 & 1.208 & 5.255 \\
     UNet &UNet & UNet	& \textbf{0.851}& 0.781	& 3.957	& 0.900	& 1.401	& 6.394& 0.876 & 1.091& 5.175 \\
     SUNet& SUNet & SUNet & 0.820 & 0.966 & 4.336 & 0.920 & 0.433	&2.810	& 0.870& 0.699 & 3.573 \\
\hline
\end{tabular}
\end{table}

\noindent\textbf{Different combination of Transformer and CNN.} As illustrated in Table \ref{tab2}, The first row represents the combination of models proposed by the DSTCT framework, and its result is the best. While collaboratively training the SUNet-SUNet-SUNet or UNet-UNet-UNet models, the performance is relatively inferior to the UNet-SUNet-SUNet models, indicating the effectiveness of the complementary of these two models. \\
\textbf{Parameter analysis.} The parameters $\alpha$, $\beta$ and $\gamma$ are essential to control the importance of each loss in the objective function. Table \ref{tab4} lists the results obtained with different parameter settings. When $\alpha$ = 0.5, $\beta$ = 1.0, $\gamma$ = 3.0, DSTCT achieves the best DSC performance of 89.32\%. 
 \begin{table}
\center
 \caption{ Initial value analysis of the trade-off weights of the loss function. The results of the PSFH segmentation are shown.}\label{tab4}
 \begin{tabular}{p{1cm} p{1cm}| c c c| c c c| c c c} 
\hline
  \multicolumn{2}{c|}{$\gamma$} & \multicolumn{3}{c|}{1.0} & \multicolumn{3}{c|}{2.0} & \multicolumn{3}{c}{3.0} \\ 
 \cline{1-11}
      $\alpha$ & $\beta$ & DSC$\uparrow$   & ASD$\downarrow$   & $HD_{95}\downarrow$  & DSC$\uparrow$   & ASD$\downarrow$   & $HD_{95}\downarrow$  & DSC$\uparrow$   & ASD$\downarrow$   & $HD_{95}\downarrow$    \\
 \hline
%  0.5	&0.5&	0.8698 & 0.8011	& 4.0662 & 0.8911 & 0.4731 & 2.7920 & 0.8833 & 0.5191 &2.7588 \\
% 1.0& 0.5& 0.8776 	&0.7289	 &3.8128	&0.8916	&0.5047 &2.7987	&0.8917 & 0.4857 & 2.8066 \\2.0& 0.5& 0.8623	& 0.9954  & 4.9167 & 0.8880	&0.5700 & 2.9571 	& 0.8907 	& 0.5134 & 2.8287 \\
 0.5& 1.0&  0.8701	&0.8985 &4.1565 &0.8917 &0.4741 &2.7726 & \textbf{0.8932}	&0.4659 & \textbf{2.7405} \\
 1.0	&1.0	&0.8671 &0.7914	&4.5107	&{0.8923} &0.4979 & 2.7862 	&0.8912	& 0.4867	&2.7949 \\
 2.0 & 1.0	&0.8829 & 0.6551 & 3.2419  & 0.8903	& 0.5501 & 2.8618	& 0.8931 	& 0.5049& 2.7588 \\
 0.5& 2.0	& 0.8732& 0.7037 & 3.8760 	& 0.8890& \textbf{0.4487} & 2.7636 	&0.8839 & 0.4840 	& 2.9388 \\
%  1.0 & 2.0	& 0.8609 & 0.9144 & 4.8015 	& 0.8907 & 0.5232&	2.8397 	&0.8918 & 0.4759	&2.7603 \\
% 2.0	&2.0	&0.8743	&0.7921 	&3.6714 	& 0.8893&	0.5520 	&2.9323 	&0.8909 	&0.5212 	&2.8218 \\
 \hline
 \end{tabular}
 \end{table}
\section{Conclusion}
In this study, we introduce a novel framework termed Dual-Student and Teacher Combining CNN and Transformer (DSTCT), designed to leverage the distinct inherent characteristics of CNN and ViT models through a synergistic training approach. The DSTCT framework intricately combines three key components: Consistency Learning with Soft Pseudo Labels (CLS), minimizing Classifier Determinacy Discrepancy (CDD), and Consistency Regularization (CR) through mean teacher architecture. This comprehensive integration aims to bolster the generalization capabilities of CNNs. Empirical evaluations conducted on a widely recognized benchmark indicate that the DSTCT framework significantly enhances the performance of CNN architecture, outstripping competing state-of-the-art methods by a substantial margin. Furthermore, this study catalyzes advancing the application of Transformer models within the realm of semi-supervised image segmentation tasks, encouraging further research and development in this promising area. The PSFHS Challenge of MICCAI 2023 and the IUGC Challenge of MICCAI 2024 are available at https://ps-fh-aop-2023.grand-challenge.org/ and https://codalab.lisn.upsaclay.fr/competitions/18413, respectively.

\section*{Acknowledgement}
Supported in part by the Natural Science Foundation of Guangdong Province under Grant 2024A1515011886 and 2023A1515012833, the Guangzhou Science and Technology Planning Project under Grant 2023B03J1297, 2024B03J1283 and2024B03J1289, Key Research and Development Program of Guangxi Province under Project No. 2023AB22074, the National Natural Science Foundation of China under Grant 61901192, and the China Scholarship Council 202206785002.
\bibliographystyle{splncs04}
\bibliography{Paper-2798}

\end{document}